\documentclass[lettersize,journal]{IEEEtran}
\usepackage{amsmath,amsfonts}
\usepackage{algorithmic}
\usepackage{algorithm}
\usepackage{array}
\usepackage[caption=false,font=normalsize,labelfont=sf,textfont=sf]{subfig}
\usepackage{textcomp}
\usepackage{stfloats}
\usepackage{url}
\usepackage{verbatim}
\usepackage{graphicx}
\usepackage{cite}
\usepackage{enumitem}
\usepackage{tabularx}
\usepackage{makecell} % 表格内换行
\usepackage[table,xcdraw]{xcolor}
\usepackage{booktabs}
\usepackage{multirow}
\usepackage{pifont}

\begin{document}
	
	\title{FUPareto: Bridging the Forgetting-Utility Gap in Federated Unlearning via Pareto Augmented Optimization}
	
	\author{Zeyan Wang, Zhengmao Liu, Yongxin Cai, Chi Li, Xiaoying Tang, Jingchao Chen, Zibin Pan* \thanks{Zibin Pan is the corresponding author}, Jing Qiu
		% <-this % stops a space
		% <-this % stops a space
		%\thanks{Manuscript received April 19, 2021; revised August 16, 2021.}}
}

\markboth{Journal of \LaTeX\ Class Files,~Vol.~14, No.~8, August~2021}%
{Shell \MakeLowercase{\textit{et al.}}: A Sample Article Using IEEEtran.cls for IEEE Journals}

\IEEEpubid{0000--0000/00\$00.00~\copyright~2021 IEEE}

\maketitle

\begin{abstract}
Federated Unlearning (FU) aims to efficiently remove the influence of specific client data from a federated model while preserving utility for the remaining clients. However, three key challenges remain: (1) existing unlearning objectives often compromise model utility or increase vulnerability to Membership Inference Attacks (MIA); (2) there is a persistent conflict between forgetting and utility, where further unlearning inevitably harms retained performance; and (3) support for concurrent multi-client unlearning is poor, as gradient conflicts among clients degrade the quality of forgetting. To address these issues, we propose FUPareto, an efficient unlearning framework via Pareto-augmented optimization. We first introduce the Minimum Boundary Shift (MBS) Loss, which enforces unlearning by suppressing the target class logit below the highest non-target class logit; this can improve the unlearning efficiency and mitigate MIA risks. During the unlearning process, FUPareto performs Pareto improvement steps to preserve model utility and executes Pareto expansion to guarantee forgetting. Specifically, during Pareto expansion, the framework integrates a Null-Space Projected Multiple Gradient Descent Algorithm (MGDA) to decouple gradient conflicts. This enables effective, fair, and concurrent unlearning for multiple clients while minimizing utility degradation. Extensive experiments across diverse scenarios demonstrate that FUPareto consistently outperforms state-of-the-art FU methods in both unlearning efficacy and retained utility.
\end{abstract}

%\begin{IEEEkeywords}
%	Federated Unlearning, privacy, Pareto Optimization
%\end{IEEEkeywords}

\section{Introduction}
Federated Learning (FL) \cite{mcmahan2017communication,cheng2020federated} enables distributed clients to collaboratively train a shared model while retaining raw data locally, establishing it as a prevalent paradigm for large-scale privacy-preserving learning. However, despite this decentralized design, recent studies demonstrate that clients’ gradient updates and parameter contributions can still implicitly leak sensitive information regarding their local training data \cite{de2021impact, zhao2023federated}. This vulnerability necessitates the implementation of the "right to be forgotten" within FL systems, as mandated by stringent data protection regulations such as the European Union’s General Data Protection Regulation (GDPR) \cite{voigt2017eu} and the California Consumer Privacy Act (CCPA) \cite{harding2019understanding}.
\begin{figure}[t]
	\centering
	\includegraphics[width=\linewidth]{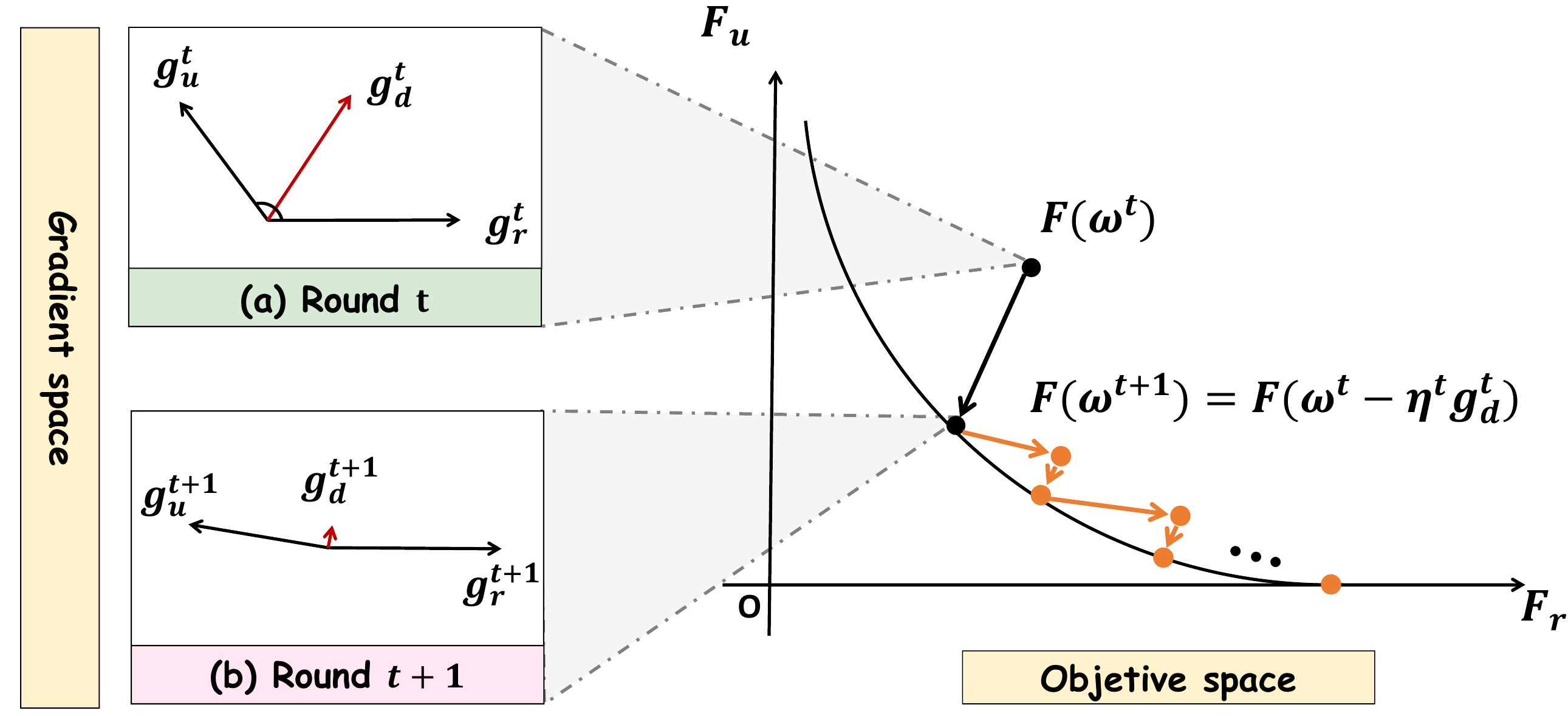}
	\caption{Gradient conflicts and the trade-off between forgetting and utility.
		Left: Gradient interactions between unlearning and remaining clients. At round $t$, the unlearning gradient $g_u^t$ and the remaining gradient $g_r^t$ are not conflict (i.e., $g_u^t \cdot g_r^t > 0$) or exhibit mild conflict, yielding a common descent direction $g_d^t$ that simultaneously minimizes both objectives. Conversely, at round $t+1$, severe gradient conflict negates this direction as the model approaches a local Pareto stationary point.
		Right: Evolution in the objective space. The updated $F(w^{t+1})$ propels the model toward a local Pareto frontier, a boundary where further decreasing $F_u$ inevitably degrades $F_r$. The orange trajectory highlights the Pareto expansion in FUPareto, which temporarily deviates from the frontier to facilitate deeper unlearning.}
	
	\label{fig:example}
\end{figure}

Federated unlearning (FU) \cite{FUsurvey1,FUsurvey2} eliminates specific data influence from trained models to ensure privacy compliance, avoiding the prohibitive costs of retraining. Prior research \cite{FUsurvey3} primarily explores: (1) archiving-based approaches (e.g., FedEraser \cite{liu2021federaser}, FedKDU \cite{wu2022federated}, FedRecovery \cite{zhang2023fedrecovery}), limited by storage overheads and inexactness; (2) gradient ascent \cite{wu2022federated}, which risks catastrophic forgetting; (3) orthogonal gradient techniques (e.g., SFU \cite{li2023subspace}, FedOSD \cite{FedOSD}) that mitigate conflicts between unlearning and remaining clients; and (4) multi-objective formulations \cite{FedOSD,pan2025multi} that balance competing goals yet fail to guarantee complete influence removal.

Building on the above efforts in FU, practical FU remains challenging. We identify three critical hurdles, primarily in:

\textbf{(1) Unlearning Objective Design.} Existing objectives generally fall into three categories: 
\begin{itemize}[leftmargin=*, noitemsep, topsep=0pt] 
	\item \textbf{Logit or accuracy suppression} forces model failure on forget data, either via gradient ascent \cite{FUPGA} or by driving target probabilities to zero using unlearning cross-entropy (UCE) \cite{wu2022federated,FedOSD,FUGAS}. However, enforcing zero probability increases susceptibility to membership inference attacks (MIA) \cite{MIA}.
	
	\item \textbf{Maximizing entropy} constrains output logits toward a uniform distribution \cite{golatkar2020eternal} to mitigate MIA risks. However, this induced randomness creates severe conflicts with remaining client objectives, significantly degrading model utility (verified in Section \ref{exp_ablation}).
	
	\item \textbf{Random labeling} trains the model on unlearning samples assigned to arbitrary classes \cite{FedAU,mora2024fedunran}. This effectively injects label-flipping noise akin to data poisoning, risking contamination of the global model. \end{itemize} Consequently, designing an objective that ensures effective forgetting while preserving MIA robustness and model integrity remains an open challenge.

\textbf{(2) Conflicts between forgetting and utility.}Even with improved objectives, reconciling unlearning quality with retained data utility remains difficult. Let $F_u(\cdot)$ and $F_r(\cdot)$ denote the unlearning and remaining objectives, respectively. As shown in Fig.~\ref{fig:example}(a), early-stage gradients from unlearning and remaining clients exhibit weak conflict, allowing a common descent direction $g_d^t$ to minimize both. However, the optimization trajectory eventually converges to a local Pareto frontier (Fig.~\ref{fig:example}(b)). Here, no direction simultaneously improves both objectives, resulting in vanishing updates that cause stagnation and incomplete unlearning. On the Pareto frontier, further reducing $F_u$ inevitably degrades $F_r$. Consequently, achieving thorough unlearning without significant utility loss remains a critical challenge.

\textbf{(3) Limited support for concurrent forgetting requests.}Most FU algorithms primarily target single-client unlearning \cite{FedAU}. Extending them to concurrent requests is non-trivial due to client-level non-IID data: gradients from multiple unlearning clients often conflict both with each other and with remaining clients. This complicates ensuring per-client forgetting quality while maintaining global utility, an issue corroborated by our experiments (Section \ref{sec_evaluation_unlearning_utility}). Therefore, developing efficient methods for simultaneous multi-client unlearning remains a key hurdle for practical deployment.

To address these challenges, we propose an \textbf{FU} framework that bridges the forgetting-utility gap via \textbf{Pareto} Augmented Optimization (\textbf{FUPareto}). First, we introduce a Minimum Boundary Shift (MBS) loss designed to eliminate target contributions with minimal parameter perturbation, thereby improving unlearning efficiency and mitigating MIA risks. Second, motivated by the trade-offs shown in Fig.~\ref{fig:example}, we devise an update mechanism alternating between Pareto improvement and expansion. Improvement steps aims to achieve both unlearning and model utility preservation, which would drive the model toward a local Pareto frontier. To prevent stagnation, the expansion step temporarily deviates from this frontier using gradient projection to preserve utility, after which subsequent iterations realign the model to restore performance. As depicted by the orange arrows in Fig.~\ref{fig:example}, this strategy allows the global model to consistently deepen unlearning while maintaining proximity to the Pareto front to mitigate the model utility reduction.

Our contributions are summarized as follows:
\begin{itemize}[leftmargin=*, noitemsep, topsep=0pt]
	\item We characterize the trade-off between forgetting efficacy and model utility in FU through a multi-objective lens, formally revealing its underlying Pareto structure. 
	\item We propose a MBS loss to eliminate target data influence while mitigating risks of MIA. 
	\item We design a Pareto-augmented optimization algorithm that guides updates toward multi-client unlearning, simultaneously preserving utility and enhancing fairness. 
	\item We conduct extensive experiments across diverse FL scenarios, demonstrating that our method consistently outperforms state-of-the-art baselines in both forgetting effectiveness and model utility.
\end{itemize}

\section{Background \& Related Work}
\subsection{Federated Learning}
FL \cite{mcmahan2017communication,cheng2020federated} facilitates collaborative global model training among multiple clients while maintaining data locality, thereby preserving data privacy. In representative algorithms such as FedAvg \cite{mcmahan2017communication}, clients conduct local training on private datasets and transmit model updates to a central server. The server subsequently aggregates these updates, typically via weighted averaging, to update the global model. Specifically, local updates are derived by optimizing a client-specific objective, commonly formulated as the cross-entropy loss:
\begin{equation}
	\mathcal{L}_{\text{CE}} = -\sum_{c=1}^{C} y_{i,c} \log(p_{i,c}),
\end{equation}
where $y_{i,c}$ denotes the true label indicator of sample $i$ and $p_{i,c}$ denotes the predicted probability of sample $i$ assigned to class $c$.

\subsection{Federated Unlearning}
FU eliminates influences of specific clients' data from a global model without retraining, addressing regulations like the GDPR right to be forgotten. It has attracted significant attention in recent years. (1) Archiving-based approaches \cite{liu2021federaser,wu2022federated,zhang2023fedrecovery,FedAU} reconstruct models via replay or calibration using stored updates, yet they incur high storage costs and approximation errors. (2) Gradient ascent methods \cite{wu2022federated} reverse optimization by maximizing loss on target data, though they risk instability with unbounded losses and utility degradation. (3) Orthogonal-gradient techniques \cite{li2023subspace,FedOSD} mitigate interference by updating in subspaces orthogonal to retained clients. (4) Multi-objective strategies \cite{FedOSD,pan2025multi} explicitly balance conflicting goals but lack guarantees under non-convex dynamics and limited observability.

FU is generally categorized into sample-level, class-level, and client-level unlearning \cite{FUsurvey2}. Following the setup in \cite{FedOSD}, this work focuses on client unlearning for two primary reasons. First, it ensures a fair comparison with archiving-based baselines such as FedEraser, which depend on pre-recorded training artifacts and are thus inapplicable to sample unlearning and class unlearning, since the target unlearning data can not be identified previously during the pre-training phase. Second, sample unlearning and class unlearning can be theoretically reducible to the client unlearning. Specifically, a subset of unlearning samples targeted for removal can be encapsulated as a virtual client $u$; this reformulation allows algorithms to handle sample unlearning and class unlearning requests through the standard client unlearning mechanism.

\subsection{Multi-Objective optimization for Unlearning}
In the context of unlearning, there is often an inherent conflict between the unlearning objective and the preservation of model utility. This conflict specifically manifests as opposing gradients between the unlearning client ($g_u$) and the remaining clients ($g_r$), i.e., $g_u \cdot g_r < 0$. Multi-objective optimization (MOO) offers an effective approach to mitigate this issue. Specifically, using the Multiple Gradient Descent Algorithm (MGDA) \cite{MGDA}, it is possible to compute a common descent direction $g_d$ that satisfies both $g_u \cdot g_d > 0$ and $g_r \cdot g_d > 0$. This simultaneously enhances unlearning efficacy while preserving model utility, a paradigm adopted by works such as FUGAS \cite{FUGAS} and MOLLM \cite{pan2025multi}. However, MOO methods often converge to local Pareto-stationary points. At such points, the conflict between unlearning and model utility becomes severe, leading to optimization stagnation that hinders the complete unlearning (as illustrated in Figure \ref{fig:example}).

\textbf{Definition: Pareto Stationarity.} A parameter vector $\omega^*$ is said to be Pareto stationary if there exist non-negative coefficients $\{\alpha_i\}$ such that $\alpha_i \geq 0$ for all $i$, $\sum_i \alpha_i = 1$, and $\sum_i \alpha_i g_i = 0$.

Our method draws from the idea of MOO to achieve the goal of unlearning while preserving model utility. Differently, we first introduce a novel MBS loss designed to accelerate unlearning efficiency and fortify resistance against MIA. Subsequently, we propose a Pareto-augmented optimization strategy to prevent the model from converging to local Pareto stationary points. Collectively, this framework effectively reconciles the inherent trade-off between unlearning and model utility retention.

\begin{figure}[t]
	\centering
	\includegraphics[width=\linewidth]{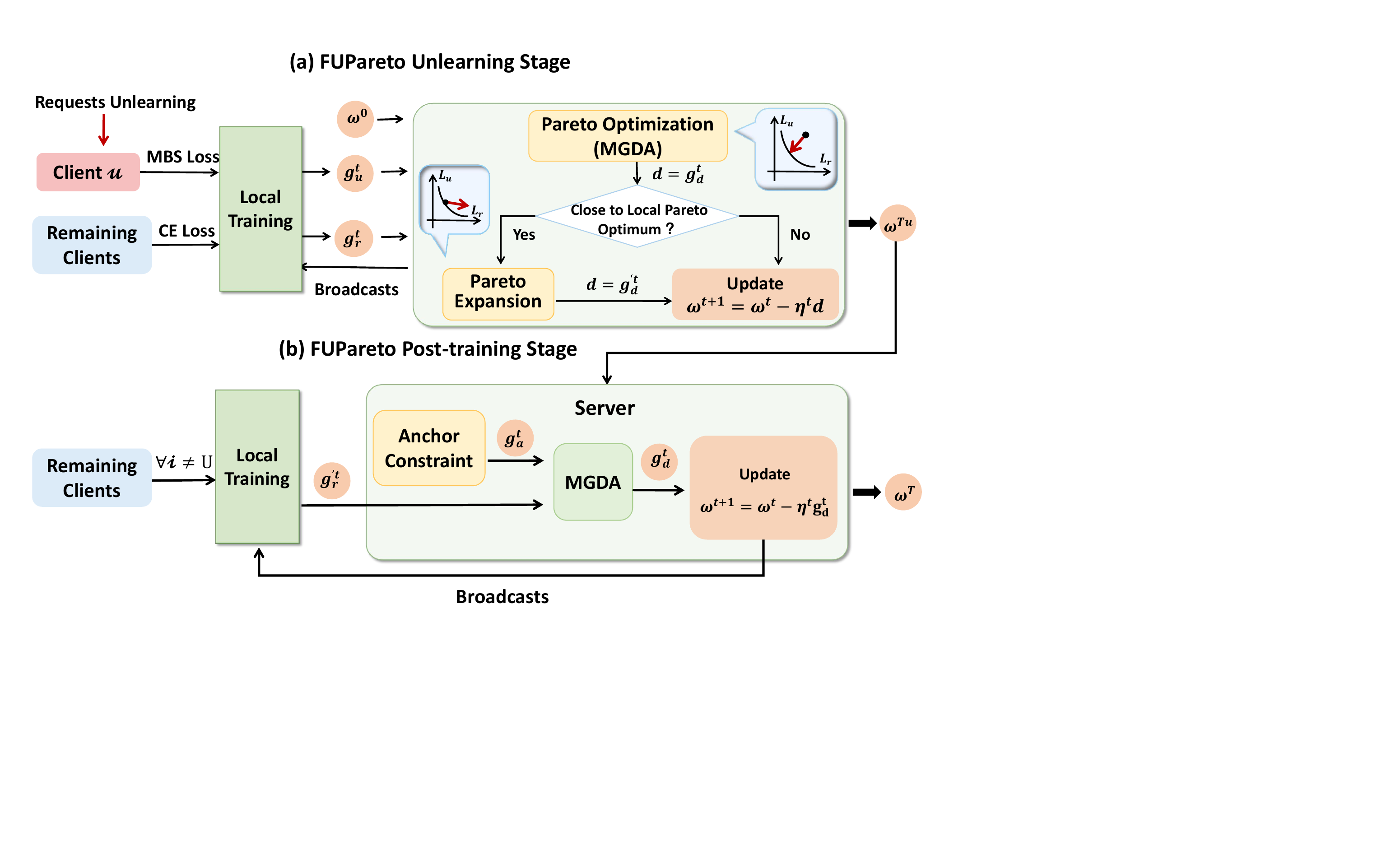}
	\caption{Overview of the FUPareto framework, illustrated with one unlearning client and one remaining client. The method comprises two stages: (a) unlearning stage, which alternates between Pareto improvement and expansion to remove target data while preserving utility, and (b) post-training with anchor-constrained multi-objective improvement to improve performance of retained clients.}
	\label{fig:kuanjia}
\end{figure}

\section{The Proposed Approach: \textnormal{FUPareto}}
\label{sec:method}

We propose FUPareto, a framework aims to achieve efficient unlearning while preserving model utility. As illustrated in Figure~\ref{fig:kuanjia} (with the full procedure detailed in Algorithm~\ref{alg:FUPareto} of the Appendix), the method initiates with an Unlearning Stage driven by a designed MBS loss. This loss is formulated to eliminate the influence of targeted data while maintaining the stability of the original decision boundaries. To better achieve unlearning requirements while preserving model utility, we incorporate Pareto augmented optimization mechanisms. Subsequently, the post-training stage employs anchor-based constraints and multi-gradient descent to realign the unlearned model with the original global model, effectively recovering performance. Additionally, a theoretical convergence analysis of the algorithm is provided in the Appendix \ref{section:B}.

The primary objective of FU is to eliminate the model's capability to correctly classify specific samples. Post-unlearning, the model must fail to predict the ground-truth label for the target data. Formally, let $p_{i,c}$ denote the predicted probability of sample $i$ for its ground-truth class $c$. Unlearning is achieved when $p_{i,c}$ ceases to be the maximum value among all classes, signifying that the sample is misclassified. Existing FU strategies suppress $p_{i,c}$ through diverse mechanisms, each with distinct drawbacks.Gradient Ascent (GA) variants and the UCE loss in FedOSD drive $p_{i,c}$ toward zero. This induces extreme output distributions that exacerbate vulnerability to Membership Inference Attacks (MIA), as empirically validated in Section \ref{sec_exp_evaluation}. KL-divergence-based methods enforce uniform distributions, which mitigates MIA risks but inherently conflicts with model utility preservation.Random-labeling approaches introduce incorrect supervision signals, effectively polluting the learned representation with out-of-distribution noise.

We indicate that forcing $p_{i,c}$ to zero or imposing uniformity is excessive. To achieve forgetting, it suffices to shift the prediction across the nearest decision boundary.Leveraging this insight, we propose the Minimum Boundary Shift (MBS) unlearning loss. This objective seeks the minimal output perturbation required to flip the prediction from the ground-truth class to the runner-up class. Let $z_{i,c}$ be the logit for the ground-truth class $c$, and $z_{i,\text{next}}$ be the maximum logit among the remaining classes:
\begin{equation}
	z_{i,\text{next}} = \max_{k \neq c} z_{i,k}.
\end{equation}

The sample-level unlearning loss is defined as:
\begin{equation}
	\mathcal{L}_{MBS} = \operatorname{ReLU}\big( z_{i,c} - z_{i,\text{next}} - \delta\big),
	\label{eq_mbs_loss}
\end{equation}
where $\delta$ is a margin ensuring separation (set to $10^{-3}$). This loss drives the model across the decision boundary. Crucially, the loss vanishes once the boundary is crossed, automatically terminating the update.Unlike prior objectives, MBS induces only the necessary shift to alter the top-1 prediction. This results in minimal parameter updates, thereby preserving utility for remaining clients. Furthermore, it avoids generating anomalous output distributions, mitigating MIA risks. Finally, the self-terminating nature of the loss ensures optimization stability.

\begin{figure}[t]
	\centering
	\includegraphics[width=0.85\linewidth]{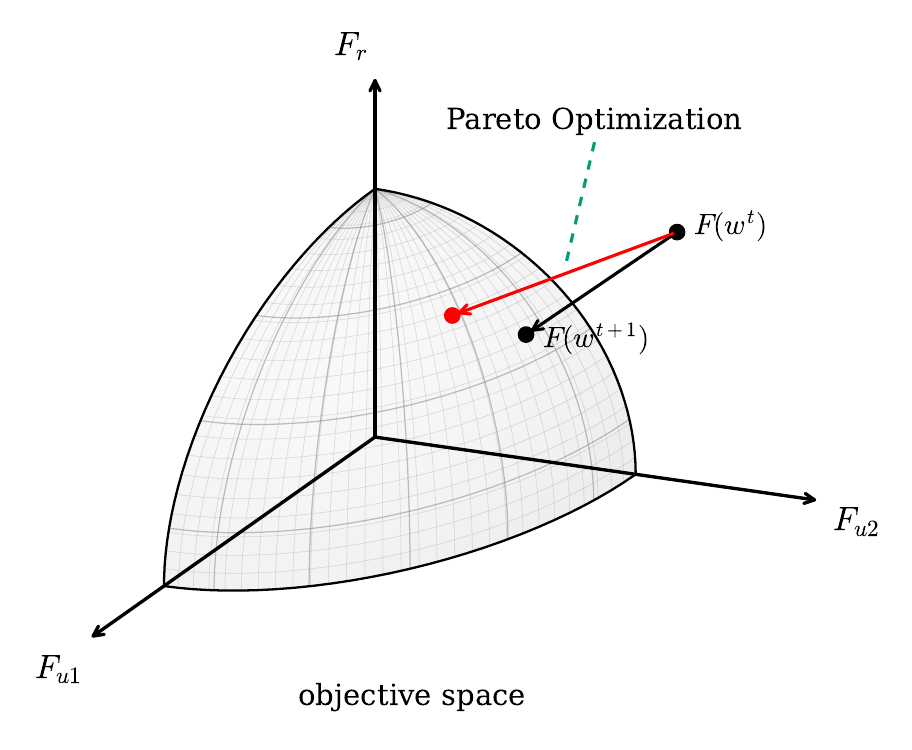}
	\caption{Illustration of the Pareto improvement step in a three-client scenario, involving unlearning clients $u_1, u_2$ and a remaining client $r$. MGDA derives a common descent direction to steer the model toward the Pareto frontier, indicated by the black arrow. Incorporating the fairness-guidance objective $g_p^t$ yields a fairer solution where the discrepancy between $F_{u_1}$ and $F_{u_2}$ is minimized, as marked by the red point.}
	\label{fig:3D}
\end{figure}

\subsection{Pareto improvement}

To preserve model utility during federated unlearning, we initially employ MGDA to steer the model toward a local Pareto frontier. Upon approaching the frontier, we execute a Pareto expansion step in the subsequent round, driving the optimization specifically toward minimizing unlearning loss. We then revert to Pareto improvement to realign the model with the frontier. This alternating mechanism effectively safeguards utility throughout the process.

% 为了在联邦遗忘过程中保持模型效用，我们首先用多梯度下降法让模型逼近局部帕累托前沿面，当模型逼近局部帕累托前沿时，我们在下一轮使用Paretttto expansion让模型继续沿着unlearning loss变小的方向移动，然后下一轮再继续用帕累托优化让模型重新回到帕累托前沿面上，实现保护模型utility。

Figure~\ref{fig:3D} visualizes the Pareto improvement step, where we model unlearning as a multi-objective optimization problem:
\begin{equation}
	\min \big\{ F_{u \in \mathcal{U}} (\omega),\; F_{r \in \mathcal{R}e}(\omega),\; F_p(\omega) \big\},
\end{equation}
where $\omega \in \mathbb{R}^n$ represents the model parameters, $n$ is the dimention. $F_{u \in \mathcal{U}}(\omega)$ are the loss of unlearning clients, $F_{r \in \mathcal{R}e}(\omega)$ is the loss of remaining clients, and $F_p(\omega)$ is a fairness-guidance objective \cite{pan2023fedmdfg} defined as
\(
F_p(\omega) = \arccos\!\left( \frac{p^\top F(\omega)}{\|F(\omega)\|} \right)
\), 
where $F(\omega)$ denotes the vector of clients' local loss. The fairness guidance vector $p \in \mathbb{R}^m$ encodes group preference: we set $p_i = 0$ for unlearning clients and $p_j = 1$ for remaining clients, namely
\(
p_i = 0,\ i \in \mathcal{U},\;
p_j = 1,\ j \in \mathcal{R}e
\).
This choice reflects the ``forget-first'' principle and can also improve unlearning performance. We verify the effect of it and the fairness-guidance objective in the ablation study (M3 and M2) in Section \ref{exp_ablation}.

To solve the multi-objective problem, we adopt MGDA to build a common descent gradient $g_d^t$ for the model update. Let $g_{u \in \mathcal{U}}^{t}$ and $g_{r \in \mathcal{R}e}^t$ be the local gradients of unlearning clients and remaining clients, respectively. Denote $g_p^t = \nabla F_p(\omega^t)$. We concatenate them to form
\(
G_t = \mathrm{concat}\!\left[\forall g_{u \in \mathcal{U}}^{t},\ \forall g_{r \in \mathcal{R}e}^t,\ g_p^t \right] \in \mathbb{R}^{n \times (m+1)}
\).
Next, we solve the following dual problem in the MGDA form:
\begin{equation}
	\lambda_t^* = \arg\min_{\substack{\lambda \in \mathbb{R}_+^{m+1} }} \left( \frac{1}{2} \lambda^\top G_t^\top G_t \lambda \right), ~ \text{s.t.}~\sum_{i=1}^{m+1} \lambda_i = 1.
\end{equation}
This is a low-dimensional quadratic program. The optimal $\lambda_t^*$ gives the best weights for the objectives.
We then compute the common descent gradient
\(
g_d^t = G_t \lambda_t^*
\),
and update the model as
\(
\omega^{t+1} = \omega^t - \eta^t g_d^t
\),
where the step size $\eta^t$ is chosen by a line search method to balance speed and stability.
This process moves the model toward a Pareto-optimal region and supports efficient and controllable FU.

\subsection{Step Size Line Search}
%改成：我们用Step Size Line Search来确定更新步长。该策略有两个好处：一是可以判断是否逼近帕累托前沿，以便在下一轮改用Pareto Expansion来继续unlearn；二是可以搜索出一个比设置的学习率更大的步长，以加速模型逼近帕累托前沿。%
%
We employ a line search strategy to determin the update step, conferring two primary advantages. First, it functions as a convergence indicator relative to the local Pareto frontier; specifically, the inability to identify a feasible step size triggers a transition to the Pareto expansion phase in the subsequent round, ensuring the unlearning process continues. Second, this approach enables the automatic selection of a learning rate exceeding the baseline, thereby accelerating convergence toward the frontier.

Formally, at round $t$, we derive a candidate direction $g_d^t$ via the Pareto improvement step and initialize the search with a magnitude of $2^s\eta$, where $s$ is a positive integer controlling the search breadth. Should the current step size fail to satisfy the Pareto improvement criterion, we iteratively halve the rate, updated as $\eta^t \leftarrow \eta^t / 2$, until a valid step is found or the lower bound $\eta_{\min} = 2^{-s}\eta$ is reached.

For each candidate step size $\eta^t$, we update parameters by
\(
\omega^{t+1} = \omega^t - \eta^t g_d^t
\),
and re-evaluate the client losses.
We accept the step size using an Armijo-style sufficient decrease rule.
For client $i$ with objective $F_i(\omega)$, we require
\begin{equation}
	F_i(\omega^{t+1}) \leq F_i(\omega^t) + \beta \eta^t \, g_i^t \cdot g_d^t, \quad \forall i \in \mathcal{S}_t.
\end{equation}
where $\beta \in (0,1)$ is the sufficient decrease factor and $\mathcal{S}_t$ is the set of clients in round $t$.
If the rule fails, we continue to reduce the step size and try again.

Failure to satisfy the Armijo condition \cite{nocedal1999numerical} within the search range $[\eta_{\min}, 2^s\eta]$ indicates the model's proximity to a local Pareto frontier, where persisting in Pareto improvement step would lead to stagnation. Consequently, the algorithm transitions to the Pareto expansion phase in the subsequent round. In this phase, the update direction is recalibrated to sustain the reduction of the unlearning objective while preserving the utility of the remaining clients.

Critically, this line search strategy incurs minimal communication overhead \cite{pan2023fedmdfg}, as clients transmit only scalar loss values rather than high-dimensional gradient vectors. From a computational perspective, the evaluation necessitates only a single forward pass, bypassing the resource-intensive backpropagation process. Since forward propagation is inherently parallelizable—and can theoretically achieve $O(1)$ time complexity on emerging memristor-based hardware \cite{cai2016low}—this approach facilitates rapid loss evaluation at the server.

\subsection{Pareto Expansion}
When the line search indicates convergence to a local Pareto frontier where conflicting objectives hinder further optimization, the algorithm transitions to the Pareto expansion phase. This phase permits a controlled deviation from the current frontier, guiding the model toward a region with lower unlearning loss while maintaining proximity to the Pareto set.

To prevent significant drift from the frontier, we employ null-space projection. Given the unlearning clients' gradient $g_u^t, u\in \mathcal{U}$ at round $t$, we project it onto the null space of the subspace spanned by the gradients of the remaining clients:

\begin{equation}
	g_u^{'t} = g_u^t - \text{proj}_{g_{r \in \mathcal{R}e}^t}(g_u^t), \forall u \in \mathcal{U}.
\end{equation}

This projection enforces strict orthogonality, satisfying $g_{u \in \mathcal{U}}^{'t} \cdot g_{r \in \mathcal{R}e}^t = 0$. Consequently, updates derived from $g_{u \in \mathcal{U}}^{'t}$ remain orthogonal to the gradients of the remaining clients. This mechanism minimizes interference with the remaining task, thereby preserving model utility without hindering subsequent optimization steps.

After we obtain the orthogonalized gradient $g_{u \in \mathcal{U}}^{'t}$, we simplify the multi-objective problem in this stage as
\begin{equation}
	\min \{ F_{u \in \mathcal{U}} (\omega), F_{p'}(\omega) \},
\end{equation}
where $F_{p'}(\omega)$ is the fairness-guidance objective, i.e., $F_p(\omega) = \arccos\!\left( \frac{p^\top F(\omega)}{\|F(\omega)\|} \right)$ with $p = \vec{1}$. Let $g_{p'}^t$ be the corresponding gradient, we concatenate the gradients $g_{u \in \mathcal{U}}^{'t}$ and $g_{p'}^t$ to form $G_t = \mathrm{concat}[\, g_{u \in \mathcal{U}}^{'t},\, g_{p'}^t \,] \in \mathbb{R}^{n \times (|\mathcal{U}|+1)}$. Following the same MGDA procedure as in Equation~(5), we obtain the common expansion direction $g_d^{'t}$. 

Finally, we update the model as $\omega^{t+1} = \omega^t - \eta^t g_d^{'t}$, where $\eta^t$ is still determined by the adaptive line search. The line search in the Pareto expansion step differs from that in Pareto improvement in two aspects. First, it only considers unlearning clients when evaluating step acceptance. Second, it searches over the reduced interval $[2^{-s} \eta,\; \eta\,]$, starting from $\eta$ rather than the enlarged initial step size $2^s \eta^t$ used in Pareto improvement. This avoids overly large updates that could push the model far from the Pareto frontier and severely harm the utility of remaining clients. 

This stage implements Pareto expansion under multiple objectives. It allows the model to keep improving after reaching a local Pareto frontier, by moving in a direction that reduces the unlearning loss while staying close to the frontier.

\begin{table*}[t]
	\centering
	\caption{ASR and retained-client accuracy (R-Acc, with standard deviation) on FMNIST, CIFAR-100, and GTSRB with 20 clients under single-unlearning client (CN=1) and multi-unlearning client (CN=5) unlearning. The row $w_0$ denotes the initial model. Results without $^\dagger$ are from the unlearning stage, and those with $^\dagger$ are after post-training.}
	\label{tab:table1}
	\tiny
	\resizebox{\textwidth}{!}{ % <<< 关键就在这一行
		\renewcommand{\arraystretch}{1}
		\begin{tabular}{lcccccccccccc}
			\toprule
			\multirow{3}{*}{\textbf{Method}}
			& \multicolumn{4}{c}{\textbf{FMNIST}} 
			& \multicolumn{4}{c}{\textbf{CIFAR-100}} 
			& \multicolumn{4}{c}{\textbf{GTSRB}} \\
			\cmidrule(lr){2-5} \cmidrule(lr){6-9} \cmidrule(lr){10-13}
			& \multicolumn{2}{c}{\textbf{CN=1}} & \multicolumn{2}{c}{\textbf{CN=5}}
			& \multicolumn{2}{c}{\textbf{CN=1}} & \multicolumn{2}{c}{\textbf{CN=5}}
			& \multicolumn{2}{c}{\textbf{CN=1}} & \multicolumn{2}{c}{\textbf{CN=5}} \\
			\cmidrule(lr){2-3} \cmidrule(lr){4-5}
			\cmidrule(lr){6-7} \cmidrule(lr){8-9}
			\cmidrule(lr){10-11} \cmidrule(lr){12-13}
			& \textbf{ASR} & \textbf{R-Acc} & \textbf{ASR} & \textbf{R-Acc} 
			& \textbf{ASR} & \textbf{R-Acc} & \textbf{ASR} & \textbf{R-Acc} 
			& \textbf{ASR} & \textbf{R-Acc} & \textbf{ASR} & \textbf{R-Acc} \\
			\midrule
			\textbf{$\omega^0$} 
			& .921 & .831(.101) & .840 & .854(.099)
			& .517 & .460(.049) & .641 & .395(.034)
			& .751 & .988(.007) & .433 & .961(.017) \\
			
			\textbf{Retraining}
			& .001 & .770(.175) & .003 & .771(.182)
			& .005 & .328(.049) & .005 & .315(.035)
			& .004 & .861(.065) & .003 & .859(.092) \\
			
			\textbf{FedEraser}
			& .002 & .604(.239) & .002 & .629(.206)
			& .007 & .137(.030) & .009 & .143(.041)
			& .000 & .245(.104) & .020 & .245(.125) \\
			
			\textbf{FUPGA}
			& .020 & .716(.266) & .288 & .608(.277)
			& .000 & .308(.064) & .116 & .327(.070)
			& .431 & .985(.008) & .093 & .925(.075) \\
			
			\textbf{FedEditor}
			& .074 & .805(.136) & .334 & .633(.195)
			& .311 & .408(.055) & .447 & .399(.029)
			& .004 & .921(.040) & .252 & .963(.016) \\
			
			\textbf{MoDe}
			& .013 & .830(.091) & .296 & .817(.120)
			& .288 & .461(.051) & .252 & .352(.045)
			& .654 & .987(.008) & .389 & .951(.023) \\
			
			\textbf{FedAU}
			& .490 & .782(.203) & .528 & .785(.218)
			& .518 & .365(.168) & .623 & .354(.227)
			& .621 & .946(.132) & .703 & .944(.308) \\
			
			\textbf{FUGAS}
			& .010 & .838(.099) & .338 & .854(.099)
			& .335 & .460(.049) & .481 & .395(.034)
			& .649 & .988(.008) & .310 & .961(.017) \\
			
			\textbf{FedOSD}
			& .004 & .837(.103) & .034 & .861(.088)
			& .010 & .469(.054) & .056 & .432(.040)
			& .008 & .986(.009) & .047 & .953(.021) \\
			
			\textbf{FUPareto (s=1)}
			& .003 & .859(.098) & .005 & .867(.088)
			& .000 & .471(.050) & .011 & .441(.039)
			& .006 & .989(.007) & .011 & .935(.055) \\
			
			\textbf{FUPareto (s=3)}
			& .001 & .851(.091) & .005 & .884(.083)
			& .003 & .475(.048) & .018 & .460(.041)
			& .006 & .989(.007) & .017 & .940(.052) \\
			
			\textbf{FUPareto (s=5)}
			& .001 & .854(.089) & .008 & .883(.083)
			& .004 & .478(.048) & .016 & .461(.039)
			& .006 & .989(.007) & .013 & .960(.025) \\
			
			\midrule
			% --- second block in the figure (highlighted rows) ---
			\textbf{FUGAS$^\dagger$}
			& .010 & .840(.094) & .334 & .860(.093)
			& .320 & .462(.050) & .453 & .404(.035)
			& .649 & .988(.008) & .300 & .962(.016) \\
			
			\textbf{FedOSD$^\dagger$}
			& .004 & .839(.099) & .044 & .857(.090)
			& .010 & .468(.053) & .073 & .437(.043)
			& .008 & .986(.009) & .047 & .955(.021) \\
			
			\textbf{FUPareto (s=1)$^\dagger$}
			& .029 & .859(.097) & .013 & .874(.088)
			& .000 & .472(.049) & .013 & .440(.039)
			& .006 & .989(.007) & .013 & .941(.050) \\
			
			\textbf{FUPareto (s=3)$^\dagger$}
			& .005 & .850(.091) & .012 & .884(.083)
			& .002 & .477(.049) & .015 & .460(.040)
			& .007 & .989(.007) & .014 & .944(.048) \\
			
			\textbf{FUPareto (s=5)$^\dagger$}
			& .006 & .855(.088) & .012 & .883(.083)
			& .007 & .477(.049) & .016 & .458(.038)
			& .007 & .989(.007) & .011 & .961(.023) \\
			
			\bottomrule
		\end{tabular}
	}
\end{table*}

\subsection{Recovery-Oriented Post-training}
As discussed, once the model approaches the Pareto frontier, further reduction in unlearning loss inevitably compromises model utility, necessitating a post-training phase for utility recovery. During this phase, we aim to bolster generalization while preventing the model from reverting to the pre-unlearning state $\omega^0$ \cite{FedOSD}. To this end, we employ a multi-objective optimization strategy incorporating an explicit constraint within the parameter space. Specifically, alongside minimizing the loss for remaining clients $F_r(\omega)$, we introduce an Anchor Constraint to regulate the deviation between the current parameters $\omega^t$ and the original model parameters $\omega^0$.
We define the anchor direction as
\begin{equation}
	g_a^t = \frac{\omega^t - \omega^{0}}{\|\omega^t - \omega^{0}\|}.
	\label{eq_anchor_gradient}
\end{equation}
We then concatenate the remaining client gradient $g_{r \in \mathcal{R}e}^t$ and the anchor gradient $g_a^t$ to form the recovery gradient matrix
$Q_{\text{rec}} = \text{concat}\big[g_{r \in \mathcal{R}_e}^t,\, g_a^t\big]$.
We use MGDA to compute the recovery-stage common descent direction $g_{d}^t$, and update the model by
$\omega^{t+1} = \omega^t - \eta^t g_d^t$.

\section{Experiments}
\subsection{Experimental Setting}
\textbf{Evaluation Metrics.} We assess FU using Attack Success Rate (ASR)~\cite{FUPGA,FedOSD} for unlearning efficacy (lower is better), retained-client test accuracy (R-Acc) for utility (higher is better), and MIA AUC to quantify membership leakage, where AUC closer to $0.5$ indicates lower attack success (see Appendix~\ref{app:mia}).
\begin{figure*}
	\centering
	\includegraphics[width=1\linewidth]{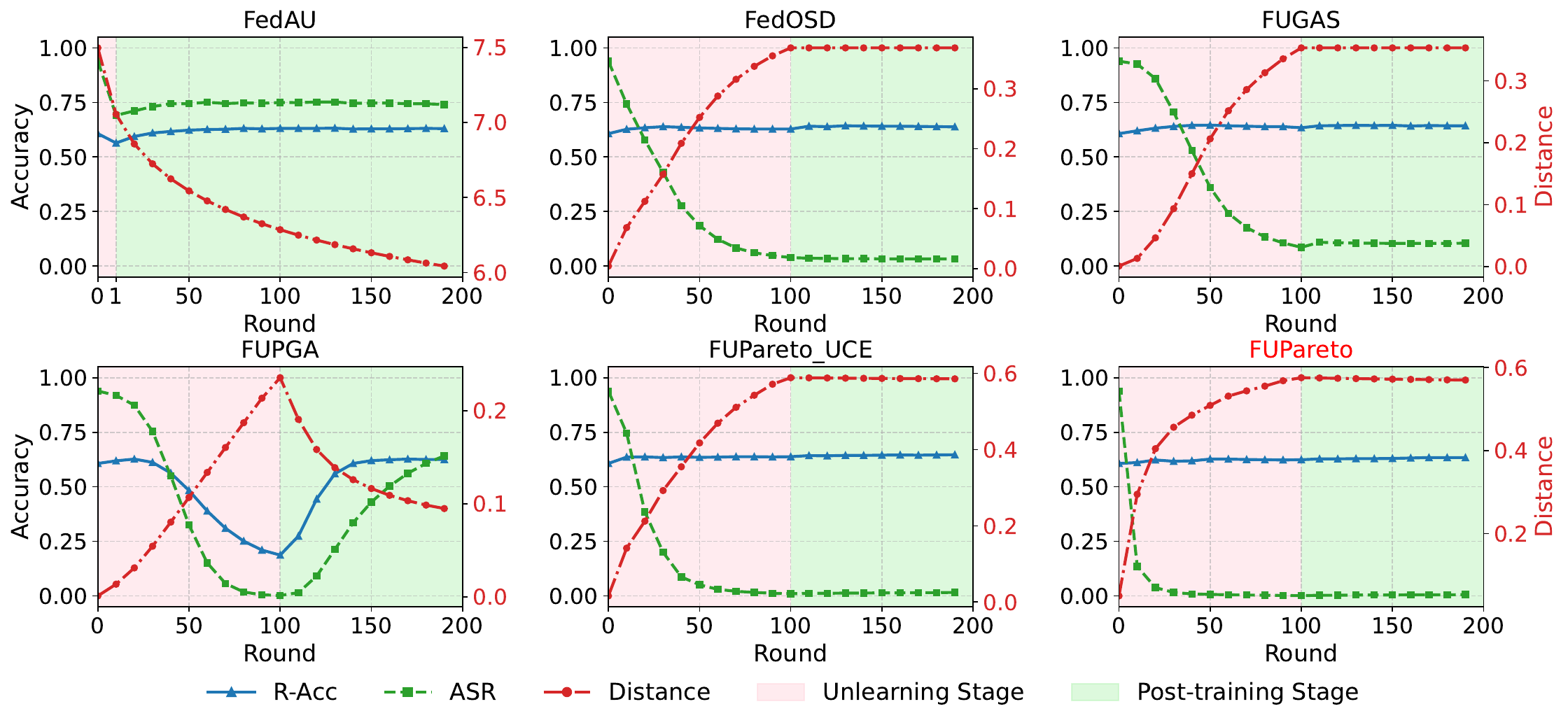}
	\caption{ASR, R-Acc, and distance from $\omega^0$ during unlearning and post-training on CIFAR-10 with 2 of 10 clients unlearning.}
	\label{fig:Convergence}
\end{figure*}

\textbf{Datasets, Models, and Settings.} Experiments are conducted on four image classification datasets: FMNIST~\cite{xiao2017fashion}, CIFAR-10, CIFAR-100~\cite{krizhevsky2009learning}, and GTSRB~\cite{stallkamp2011german}, using LeNet-5~\cite{lecun2002gradient}, CNN~\cite{lecun2002gradient}, NFResNet-18~\cite{brock2021characterizing}, and CNN, respectively. We simulate data heterogeneity with Dirichlet partitions ($\alpha=0.1, 0.5$), and also consider Pat NC=10 where each client contains all classes, yielding an approximately IID distribution. The initial model ($\omega^0$) is obtained by running FedAvg for 2,000 communication rounds. The unlearning phase lasts 100 rounds, resulting in 200 rounds in total (including the phase following training). We evaluate unlearning at different scales, from single-client to concurrent multi-client unlearning, including fixed numbers of unlearning clients (e.g., 1 or 5), proportions (e.g., 20\%), and client populations of 10, 20, and 50.

\textbf{Baselines and Hyperparameters.} We compare FUPareto with Retraining, FedEraser~\cite{liu2021federaser}, FUPGA~\cite{FUPGA}, FedEditor~\cite{yuan2025fededitor}, MoDe~\cite{zhao2023federated}, FedAU~\cite{FedAU}, FUGAS~\cite{FUGAS}, and FedOSD~\cite{FedOSD}; the initial model $\omega^0$ is reported as a reference. For all methods, we set local epochs $E=1$, use SGD with batch size 200, apply a learning rate decay of 0.999 per round, and set the step-size search parameter $s \in \{1,3,5\}$.

\subsection{Evaluation of Unlearning Effectiveness and Utility} \label{sec_evaluation_unlearning_utility}
As detailed in Table~\ref{tab:table1}, existing methods struggle to balance unlearning effectiveness and model utility in the 20-client setting. Gradient-ascent-based approaches like FUPGA sacrifice utility to reduce ASR, while archiving-based methods such as FedEraser exhibit instability and low accuracy under complex distributions. Furthermore, FedEditor, MoDe, and FUGAS fail to achieve effective forgetting in concurrent unlearning scenarios, yielding high ASRs. FedAU similarly suffers from insufficient unlearning and severe model reversion.

In contrast, FUPareto achieves a superior trade-off by consistently maintaining low ASR and high accuracy across all scenarios, demonstrating robustness in multi-client unlearning. Additional experiments in Appendix~\ref{app:sequential} confirm that while prior methods require computationally expensive sequential unlearning to reach acceptable results, FUPareto delivers optimal performance directly, significantly reducing communication and computation overhead.

\begin{figure}[h!]
	\centering
	\includegraphics[width=0.90\columnwidth]{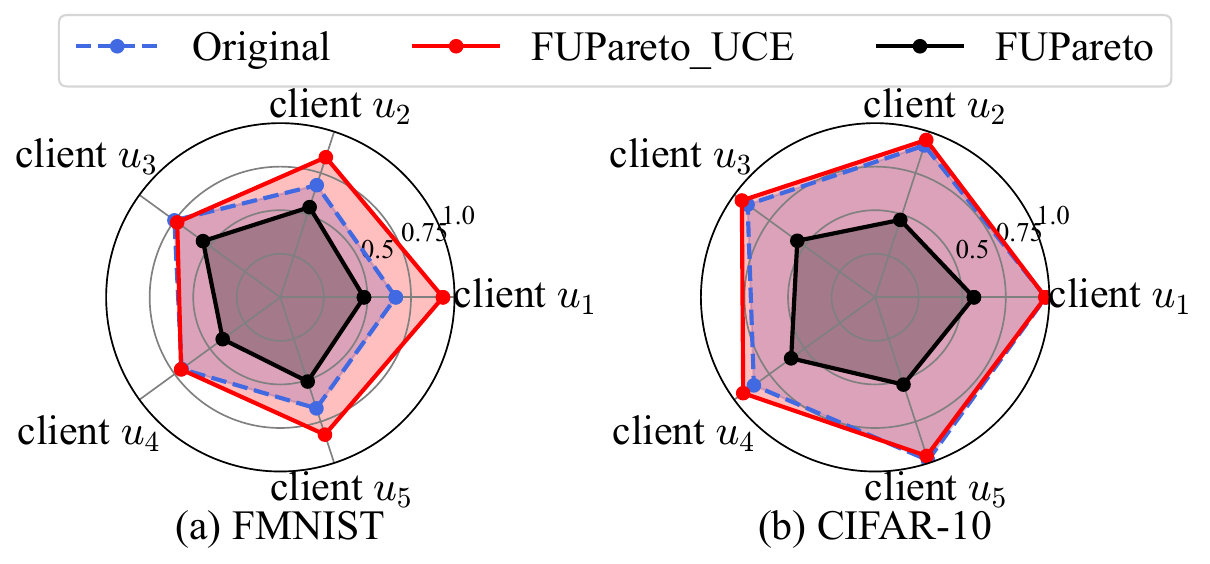}
	\caption{AUC of each unlearning client on (a) FMNIST and (b) CIFAR-10.}
	\label{fig:radar}
\end{figure}
We further investigate scalability and robustness regarding client population size and data heterogeneity. As client numbers grow or data distributions become more heterogeneous, most baselines suffer from significant degradation in either unlearning efficacy or utility stability, primarily due to exacerbated gradient conflicts. In contrast, FUPareto maintains consistent unlearning performance and high retained accuracy across varying scales and distributions, including IID and Dirichlet non-IID settings. Detailed quantitative results and runtime analysis are provided in Appendix~\ref{app:robustness} and Appendix~\ref{app:runtime}, respectively.

Finally, we observe that FUPareto is not sensitive to the step-size parameter $s$ within a reasonable range. While a small value may trigger premature Pareto expansion and affect utility preservation, moderate settings yield similar ASR and R-Acc with lower overhead. Based on this observation, we adopt $s=3$ in all experiments.

\subsection{Convergence Analysis.}
Figure~\ref{fig:Convergence} illustrates the convergence behavior on CIFAR-10 with 10 clients (2 unlearning targets). While FedAU maintains high R-Acc, it suffers from slow convergence and persistently high ASR. FedOSD converges rapidly but exhibits significant utility degradation. FUGAS tends to stagnate at local Pareto frontiers, hindering further unlearning progress. Similarly, FUPGA experiences model reversion during post-training, evidenced by rising ASR and partial information recovery. Conversely, FUPareto achieves rapid R-Acc convergence and maintains consistently low ASR. Its adaptive post-training mechanism effectively mitigates model reversion, ensuring stable and efficient unlearning.

Furthermore, we evaluate the impact of the loss function by replacing our MBS loss with the unlearning cross-entropy (UCE) loss~\cite{FedOSD}. The slower ASR reduction observed with UCE suggests that rigidly forcing $p_{i,c}$ toward zero compromises unlearning efficiency compared to our approach.

\subsection{MBS Loss Mitigates MIA Risk}\label{sec_exp_evaluation}
We further validate the privacy benefits of the proposed MBS loss against Membership Inference Attacks (MIA) \cite{shokri2017membership}. An attack model is trained to distinguish whether a specific sample originates from an unlearning client, quantifying the risk via the Area Under the Curve (AUC) metric (where 0.5 implies ideal indistinguishability). As visualized in Figure \ref{fig:radar}, the pre-unlearning model exhibits high AUC scores across all five target clients. Post-unlearning, FUPareto significantly reduces this risk, driving AUC values toward 0.5. This indistinguishability stems from the design of MBS loss, which avoids excessive suppression of target logits.

To isolate the source of this improvement, we conduct an ablation study by substituting MBS with UCE loss (denoted as FUPareto\_UCE). This modification leads to a substantial spike in AUC, indicating heightened privacy leakage. The vulnerability arises because UCE loss minimizes target logits without a lower bound; this extreme behavior creates a distinct statistical signature that attackers can easily exploit.

\subsection{Ablation Study} \label{exp_ablation}

We present an ablation study to examine the contribution of individual components in FUPareto, with results summarized in Table \ref{tab:ablation}.

\begin{itemize}[leftmargin=*, itemsep=0pt, topsep=2pt] 
	\item \textbf{M1 (Naïve Update)} bypasses Pareto optimization and utilizes the weighted sum of unlearning clients' gradients to update the model. While achieving low ASR, it suffers from catastrophic utility degradation and high variance. 
	\item \textbf{M2 (No Fairness)} removes the fairness objective, compromising both unlearning efficacy and stability, particularly on complex datasets like CIFAR-100. 
	\item \textbf{M3 (Uniform Preference)} discards the "forget-first" priority. It fails to unlearn effectively on CIFAR-100, validating the need for targeted objective prioritization. 
	\item \textbf{M4 (Unguided Expansion)} excludes fairness guidance during expansion, which effectively reduces ASR but consistently harms R-Acc. 
	\item \textbf{M5 (No Projection)} omits null-space projection, causing significant interference with retained clients and reducing model utility. 
	\item \textbf{M6 (Aggressive Step)} enlarges the expansion step size, destabilizing the training process and degrading R-Acc. 
	\item \textbf{M7 (UCE Loss) \& M8 (KL Divergence)} replace MBS loss with other unlearning losses. M7 induces over-forgetting, while M8 severely destroys output structure and utility. 
\end{itemize}

Collectively, these ablations confirm that each component is indispensable. Removing Pareto optimization, fairness guidance, or null-space projection leads to instability and utility loss, while replacing the bounded MBS loss with aggressive alternatives causes over-forgetting. FUPareto effectively integrates these modules to achieve a robust trade-off between unlearning and utility.

\begin{table}
	\centering
	\caption{ASR and retained-client accuracy (R-Acc, with standard deviation) of ablation variants on FMNIST, CIFAR-10, and CIFAR-100 with 10 clients, where 3 clients request unlearning.}
	
	\label{tab:ablation}
	\scalebox{0.73}{
		\renewcommand{\arraystretch}{1.1}
		\begin{tabular}{lcccccc}
			\toprule
			\multirow{2}{*}{\textbf{Method}} 
			& \multicolumn{2}{c}{\textbf{FMNIST}} 
			& \multicolumn{2}{c}{\textbf{CIFAR-10}} 
			& \multicolumn{2}{c}{\textbf{CIFAR-100}} \\
			\cmidrule(lr){2-3} \cmidrule(lr){4-5} \cmidrule(lr){6-7}
			& \textbf{ASR} & \textbf{R-Acc} 
			& \textbf{ASR} & \textbf{R-Acc} 
			& \textbf{ASR} & \textbf{R-Acc} \\
			\midrule
			$\omega^0$ 
			& .804 & .865 (.084) 
			& .898 & .606 (.094) 
			& .766 & .419 (.025) \\
			
			FUPareto 
			& \textbf{.010} & \textbf{.879 (.096)} 
			& \textbf{.021} & \textbf{.639 (.090)} 
			& \textbf{.058} & \textbf{.489 (.039)} \\
			
			M1 
			& .000 & .557 (.349) 
			& .001 & .335 (.225) 
			& .000 & .168 (.080) \\
			
			M2 
			& .002 & .848 (.104) 
			& .002 & .583 (.112) 
			& .186 & .491 (.043) \\
			
			M3 
			& .012 & .882 (.085) 
			& .024 & .642 (.090) 
			& .111 & .492 (.039) \\
			
			M4 
			& .001 & .847 (.098) 
			& .001 & .592 (.109) 
			& .052 & .481 (.040) \\
			
			M5 
			& .016 & .650 (.069) 
			& .000 & .436 (.193) 
			& .085 & .350 (.039) \\
			
			M6 
			& .000 & .826 (.152) 
			& .000 & .510 (.129) 
			& .085 & .418 (.089) \\
			
			M7 
			& .001 & .839 (.123) 
			& .001 & .608 (.135) 
			& .066 & .420 (.089) \\
			
			M8 
			& .006 & .758 (.097) 
			& .012 & .542 (.194) 
			& .017 & .262 (.098) \\
			\bottomrule
		\end{tabular}
	}
\end{table}
\subsection{Conclusion and Future Work}
This paper uncovers the intrinsic Pareto conflict between unlearning effectiveness and retained model utility in federated unlearning. We demonstrate that existing multi-objective methods often stagnate at local Pareto stationary points, particularly under non-IID and concurrent unlearning settings, leading to incomplete forgetting or utility degradation. To address this, we propose FUPareto, a Pareto-expansion-based federated unlearning framework. Specifically, we introduce an MBS unlearning loss to mitigate MIA risks and ensure non-polluting forgetting. By alternating between Pareto improvement and expansion, our approach advances the unlearning objective while maintaining proximity to the Pareto frontier, thereby facilitating scalable and fair concurrent unlearning. Extensive experiments demonstrate that FUPareto outperforms state-of-the-art methods in both unlearning effectiveness and retained utility. Future work will extend the framework to diverse loss functions and tasks, incorporate formal privacy guarantees, and explore adaptive Pareto expansion strategies for large-scale systems.

\bibliographystyle{IEEEtran}
\bibliography{reference}
%

% your own custom macro containing the $\backslash${\tt{includegraphics}} command to make things
%\bf{If you include a photo:}\vspace{-33pt}
%\begin{IEEEbiography}[{\includegraphics[width=1in,height=1.25in,clip,keepaspectratio]{fig1}}]{Michael Shell}
%Use $\backslash${\tt{begin\{IEEEbiography\}}} and then for the 1st argument use $\backslash${\tt{includegraphics}} to declare and link the author photo.
%Use the author name as the 3rd argument followed by the biography text.
%\end{IEEEbiography}
%
%\vfill

\end{document}